\documentclass[final]{cvpr}
\newcommand{\ignore}[1]{}
\usepackage{times}
\usepackage{epsfig}

\usepackage{microtype}
\usepackage[normalem]{ulem}
\usepackage{xcolor}
\usepackage{graphicx}
\usepackage{subfig}
\usepackage{booktabs} 
\usepackage{enumitem}
\usepackage{xargs}
\usepackage{algorithm}
\usepackage[noend]{algpseudocode}
\usepackage{eqparbox}
\usepackage{multirow}
\usepackage{makecell}
\usepackage{amsmath,amssymb,mathtools}
\usepackage{dsfont}
\usepackage{amsthm}
\theoremstyle{definition}
\usepackage{footmisc}

\usepackage{stackengine}
\usepackage{textcomp}

\usepackage{color,soul}

\DeclareMathOperator*{\argmax}{argmax}

\newcommand\redsout{\bgroup\markoverwith{\textcolor{red}{\rule[0.5ex]{2pt}{0.4pt}}}\ULon}
\newcommand{\NA}{\textbf{N/A}}

\usepackage[pagebackref=true,breaklinks=true,colorlinks,bookmarks=false]{hyperref}

\def\cvprPaperID{6361} 
\def\confYear{CVPR 2021}
\begin{document}
\title{Localized Uncertainty Attacks}

\author{Ousmane Amadou Dia\\
Facebook\\
{\tt\small ousamdia@fb.com}
\and
Theofanis Karaletsos\\
Facebook\\
{\tt\small theokara@fb.com}
\and
Caner Hazirbas\\
Facebook\\
{\tt\small hazirbas@fb.com}
\and
Cristian Canton Ferrer\\
Facebook\\
{\tt\small ccanton@fb.com}
\and
Ilknur Kaynar Kabul\\
Facebook\\
{\tt\small ilknurkabul@fb.com}
\and
Erik Meijer\\
Facebook\\
{\tt\small erikm@fb.com}
}

\maketitle

\begin{abstract}
The susceptibility of deep learning models to adversarial perturbations has stirred renewed attention in adversarial examples resulting in a number of attacks. However, most of these attacks fail to encompass a large spectrum of adversarial perturbations that are imperceptible to humans. In this paper, we present localized uncertainty attacks, a novel class of threat models against deterministic and stochastic classifiers. Under this threat model, we create adversarial examples by perturbing only regions in the inputs where a classifier is uncertain. To find such regions, we utilize the predictive uncertainty of the classifier when the classifier is stochastic or, we learn a surrogate model to amortize the uncertainty when it is deterministic. Unlike $\ell_p$ ball or functional attacks which perturb inputs indiscriminately, our targeted changes can be less perceptible. When considered under our threat model, these attacks still produce strong adversarial examples; with the examples retaining a greater degree of similarity with the inputs.
\end{abstract}

\section{Introduction}
\label{introduction}
\noindent In this paper, we present \textit{localized uncertainty attacks}, a novel class of threat models for creating adversarial examples against deep learning models. Under this threat model, adversarial examples are generated via \textit{replacement} by localizing regions in the original inputs where a classifier is \textit{uncertain or makes unconstrained extrapolations}. By ``replacement'', we mean that instead of perturbing indiscriminately an input as standard threat models intend to, \textit{only its select regions or points} are subject to adversarial changes. We motivate our attack model in more details in  Appendix~\ref{sec:motivations}.
\vskip 0.1cm

\noindent Unlike standard threat models~\cite{DBLP:journals/corr/abs-1802-00420,DBLP:journals/corr/abs-1712-02779,DBLP:journals/corr/CarliniW16a}, which perturb indiscriminately every input, our attack model focuses only on select few pixels; resulting thus in improved imperceptibility while maintaining similar attack success rates. Our perturbation hinges on quantifying the uncertainty associated with the predictions of the target classifiers on specific regions of the inputs. For deterministic classifiers, we train surrogate uncertainty models to get amortized uncertainty estimates. To the authors' knowledge, there is no or limited work on using uncertainty to attack both \textit{deterministic and stochastic} deep learning models. Our threat model can also be coupled with standard threat models to achieve even stronger attacks.
\vskip 0.1cm

\section{Uncertainty Threat Model}
\noindent We propose localized uncertainty attacks, a novel class of threat models for creating adversarial examples against deep learning models. Under this threat model, adversarial examples are generated via replacement by localizing regions in the input images where a classifier $g_\theta$ is poorly constrained. \vskip 0.1cm

\begin{figure*}[thb]
\vskip -0.1in
\begin{center}
    \includegraphics[width=16cm]{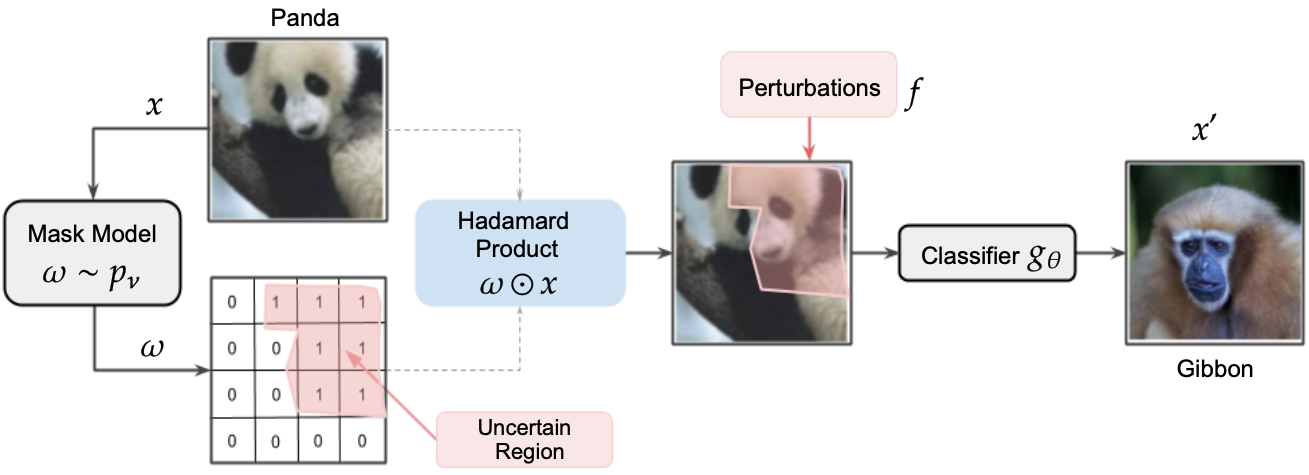}
\end{center}
\caption{\small{Given the image of a panda as input, the mask model $p_{\nu}$ learns to localize regions in the input where the classifier $g_{\theta}$ is uncertain (or makes unconstrained extrapolations) by outputting a binary mask $\boldsymbol{\omega}$. The pink region in the mask denotes a region in the input where $g_{\theta}$ is poorly constrained. When attacking the input, only the pixels lying in that region will be affected by the perturbation function $f$.}}
\label{fig:arch}
\end{figure*}

\noindent \textbf{Framework.} In Figure~\ref{fig:arch}, we provide a high-level depiction of our framework. Essentially, our framework consists of a mask model which, when given as input a raw image $\boldsymbol{x}$, learns to locate regions within $\boldsymbol{x}$ where $g_{\theta}$ makes unconstrained extrapolations. The mask model learns a distribution $p_{\nu}(\boldsymbol{x})$ over binary masks; with each mask $\boldsymbol{\omega}$ delineating regions in $\boldsymbol{x}$ to perturb so as to induce misclassification. \vskip 0.10cm

\noindent For any given input image $\boldsymbol{x}$, each mask $\boldsymbol{\omega}=(\omega_1, ..., \omega_n)$ is of similar shape as $\boldsymbol{x}=(x_1, ..., x_n)$ and maintains a one-to-one coordinate mapping with $\boldsymbol{x}$. Ideally, we want a mask $\boldsymbol{\omega}\sim p_{\nu}(\boldsymbol{x})$ which, when applied to $\boldsymbol{x}$ (or $\boldsymbol{\omega}\odot\boldsymbol{x}$), maximizes the predictive uncertainty of $g_{\theta}$. What this means is that the training of the mask model is based upon maximizing the predictive uncertainty of $g_{\theta}$. We describe in Section~\ref{sec:uncertainty} how we estimate such uncertainty. Finally, we want the perturbations we apply to $\boldsymbol{\omega}\odot\boldsymbol{x}$ to become adversarial. Understandably, we need to regularize the mask model so that the masks it outputs are neither full, i.e.; $\omega_1=...=\omega_n=1$, nor completely sparse, i.e.; $\omega_1=...=\omega_n=0$.
\vskip 0.05in

\noindent \textbf{Uncertainty Threat Model.}
We consider a \textit{white-box scenario} where we want to craft \textit{strong} and \textit{visually imperceivable} adversarial examples by altering as few pixels as possible. We generate adversarial examples by solving:
\begin{equation}
    \begin{gathered}
        \boldsymbol{x'} = \argmax_{\boldsymbol{x'}\in\mathcal{S}(f(\boldsymbol{x},\, \boldsymbol{\omega}))} \boldsymbol{\mathcal{L}_{\text{adv}}}(\boldsymbol{x'}, \boldsymbol{y}), \text{ where } \boldsymbol{\omega}\sim p_{\nu}(\boldsymbol{x}) \text{ and}\\
        f(x_i,\,\omega_i) = 
        \begin{cases}
            f(x_i) + \delta_i,& \text{if }\omega_i = 1\\
            x_i,              & \text{otherwise}
        \end{cases}
        \text{ for }\, i=1, ..., n.
    \end{gathered}
    \label{eq:unc}
\end{equation}
Here, $\mathcal{S}(\boldsymbol{x})$ denotes the set of permissible perturbations the attacker can inject to $\boldsymbol{x}$ to induce misclassification from $g_{\theta}$. A common choice for $\mathcal{S}(\boldsymbol{x})$ is an $\ell_p$ ball of radius $\epsilon$ centered on $\boldsymbol{x}$, defined as: $\mathcal{S}(\boldsymbol{x})=\{\boldsymbol{x}\, +\, \boldsymbol{\delta}: \|\boldsymbol{\delta}\|_p\leq \epsilon\}$, for some norm $\|.\|_p$ where $p\in\{0, 1, 2, \infty\}$. $\boldsymbol{\mathcal{L}_{\text{adv}}}$ denotes the adversarial attack loss. If $g_\theta$ is deterministic, then  $\boldsymbol{\mathcal{L}_{\text{adv}}}(\boldsymbol{x'}, \boldsymbol{y}) = \ell(\boldsymbol{x'}, \boldsymbol{y}; \theta)$ else, $\boldsymbol{\mathcal{L}_{\text{adv}}}(\boldsymbol{x'}, \boldsymbol{y}) = \mathbb{E}_{\theta\sim q_{\theta}} \ell(\boldsymbol{x'}, \boldsymbol{y}; \theta)$ where $q(\theta|\mathcal{D})$ is the predictive posterior over the weights $\theta$. 
\vskip 0.15cm

\noindent Our formalism extends gracefully the additive and functional threat models. By defining $f$ as the identity function $f\colon x\mapsto x$ where $x\in\boldsymbol{\mathcal{X}}$, we recover the additive threat models. By setting $\delta_i=0$ for all $i$, we get the functional threat models. Since our threat model extends aforementioned threat models, it inherits their strengths without sacrificing imperceptibility as only few points of $\boldsymbol{x}$ are perturbed.
\vskip 0.15cm

\noindent In what follows, we consider separately the $\ell_p$ attacks under our threat model (that is; we set $f$ as an identity function in Equation~\ref{eq:unc}), and the functional attacks (that is; $\delta_1=...=\delta_n=0$) under our threat model. When experimenting with the functional attacks, we use the color space transformation function proposed by~\cite{DBLP:journals/corr/abs-1906-00001}. Formally, we treat each pixel $x_i$ in the input image $\boldsymbol{x}$ as a point in a 3-dimensional RGB color space, i.e.; $x_i = (c_{i, 1}, c_{i, 2}, c_{i, 3})$ where $c_{i, *}\in[0, 1]$. Then, we use the trilinear interpolation introduced by~\cite{DBLP:journals/corr/abs-1906-00001} as the perturbation function $f$ to modify the pixels under the same imperceptibility constraints that they observe for $f$. 

\section{Localized Uncertainty Attacks}
\noindent A prevalent explanation for the existence of adversarial examples is that they lie off the manifold of natural inputs~\cite{2019arXiv190303905A,DBLP:journals/corr/abs-1811-00525}, occupying regions where the model makes unconstrained extrapolations~\cite{smith2018understanding}. Using this hypothesis as a guiding principle, we could arguably target these regions in order to produce adversarial examples. What this hinges on here is first perturbing parts of an input and measuring the distance of the input to the data manifold. Then, we can quantify the uncertainty associated with the predictions of a target model on the input. One way to measure the uncertainty of a model is to compute the entropy of its predictive distribution. 

\subsection{Uncertainty Estimation}
\label{sec:uncertainty}
\vskip 0.1cm

\noindent Taking inspiration from~\cite{smith2018understanding}, we use entropy in the probability space as a proxy for classification uncertainty. Next, we describe how we measure uncertainty. 
\vskip 0.1cm

\noindent \textbf{Entropy.} In the canonical classification setting, where the output of the classifier $g_{\theta}$ is a conditional probability distribution $p(\boldsymbol{y}|\boldsymbol{x}; \theta)$ over the discrete set of outcomes $\boldsymbol{\mathcal{Y}}$, the entropy of the predictive distribution of $g_{\theta}$ is defined by:
\begin{equation*}
    \begin{gathered}
        H\big[p(\boldsymbol{y}|\boldsymbol{x}; \theta)\big] = - \sum\limits_{\boldsymbol{y}\in\boldsymbol{\mathcal{Y}}} p(\boldsymbol{y}|\boldsymbol{x}; \theta) \log p(\boldsymbol{y}|\boldsymbol{x}; \theta).
    \end{gathered}
\end{equation*}

\noindent \textbf{Mutual Information.} According to~\cite{smith2018understanding}, a good measure of uncertainty that is able to distinguish epistemic from aleatoric examples is the mutual information between the model parameters and the data, defined as: 
\begin{equation}
    \begin{gathered}
        U_e(\theta, \boldsymbol{y}|\boldsymbol{\mathcal{D}}, \boldsymbol{x}) = \underbrace{H\big[p(\boldsymbol{y}|\boldsymbol{x}, \boldsymbol{\mathcal{D}})\big]}_{U_t(\boldsymbol{y}|\boldsymbol{x}, \boldsymbol{\mathcal{D}})} - \underbrace{\mathbb{E}_{p(\theta|\boldsymbol{\mathcal{D}})} H\big[p(\boldsymbol{y}|\boldsymbol{x}; \theta)\big]}_{U_p(\boldsymbol{y}|\boldsymbol{x}, \theta)}.
    \end{gathered}
    \label{eq:mi}
\end{equation}
The quantities $U_t$ and $U_p$ are referred to as the predictive entropy and the expected entropy. Mutual information can provide good estimates of a model epistemic uncertainty~\cite{smith2018understanding}. In the form presented above, $U_e$ is not practical as the predictive distribution $p(\boldsymbol{y}|\boldsymbol{x}, \boldsymbol{\mathcal{D}})$ is not analytically tractable. However, $p(\boldsymbol{y}|\boldsymbol{x}, \boldsymbol{\mathcal{D}})$ can be approximated using the Bayesian interpretation of dropout~\cite{JMLR:v15:srivastava14a,2015arXiv150602142G}, variational inference~\cite{10.5555/2986459.2986721}, or the Mean-Field variational approximation of~\cite{blundell2015weight}. In either cases, $U_e$ becomes readily computable using the following:
\begin{equation*}
    \begin{gathered}
        p(\boldsymbol{y}|\boldsymbol{\mathcal{D}}, \boldsymbol{x}) \approx \frac{1}{T} \sum\limits_{t=1}^T p(\boldsymbol{y}|\boldsymbol{x}, \theta_t) \text{ where } \theta_t\sim q(\theta|\boldsymbol{\mathcal{D}}).
    \end{gathered}
    \label{eq:mf}
\end{equation*}
\noindent Here, $q(\theta|\boldsymbol{\mathcal{D}})$ denotes the dropout distribution when using dropout variational~\cite{2015arXiv150602142G}, or the variational approximation to the true posterior $p(\theta|\boldsymbol{\mathcal{D}})$ when using (mean-field) variational inference~\cite{blundell2015weight,10.5555/2986459.2986721}, and $\theta_t$ are its samples. With this approximation, we can compute  $U_e$ tractably:
\begin{equation}
    \begin{split}
        U_e &\approx H\big[\frac{1}{T} \sum\limits_{t=1}^T p(\boldsymbol{y}|\boldsymbol{x}, \theta_t)\big] - \frac{1}{T} \sum\limits_{t=1}^T H \big[p(\boldsymbol{y}|\boldsymbol{x}, \theta_t)\big].
    \end{split}
    \label{eq:mi2}
\end{equation}
\noindent \textbf{Amortized Uncertainty.} Computing the mutual information  $U_e$ in Equation~\ref{eq:mi2} requires $g_{\theta}$ to be stochastic. More formally, in Equation~\ref{eq:mi2} the model weights $\theta_t$ which parameterize $g_\theta$ are considered as random variables that follow some specific distributions. Essentially, this means that $g_\theta$ outputs conditional probabilities rather than point-estimates as it is typically the case for deterministic classifiers. Unfortunately, most of the existing adversarial attacks consider deterministic classifiers. Although we are in a white-box setting where we have complete knowledge about $g_{\theta}$, we cannot modify $g_{\theta}$ to render its weights stochastic. To circumvent this issue, we learn a surrogate uncertainty model $\mathcal{C}_{\phi}$ to estimate the quantities $U_t$ and $U_p$, and compute $U_e$. We validate empirically these models in Appendix~\ref{sec:results}.
\vskip 0.1cm

\noindent Let $g_{\bar{\theta}}$ denote the embedding sub-network of $g_{\theta}$, where $\bar{\theta} \subset \theta$. Given an input $\boldsymbol{x}\in\boldsymbol{\mathcal{X}}^n$, let $h=g_{\bar{\theta}}(\boldsymbol{x})$ denote the embedding of $\boldsymbol{x}$ --- one can view $h$ as the output we get at the penultimate layer of $g_{\theta}$. We train $\mathcal{C}_{\phi}$ by sampling from the Gaussian distribution $\mathcal{N}(h, \text{diag}(\sigma\odot\sigma))$, where the standard deviation $\sigma$ is outputted by $\mathcal{C}_{\phi}$ when given $h$: $\sigma = \mathcal{C}_{\phi}(h)$. Explicitly, to train $\mathcal{C}_{\phi}$ we sample $M$ representations $\boldsymbol{z_i}\sim \mathcal{N}(h, \text{diag}(\sigma\odot\sigma))$ for every input $\boldsymbol{x}$. Then, we optimize $\mathcal{C}_{\phi}$ via cross-entropy using the predictive label distribution of each $\boldsymbol{z_i}$ defined by the multinomial distribution:
\begin{equation*}
    \begin{gathered}
        p(\boldsymbol{y}|\boldsymbol{z_i}; \phi) = \textbf{Softmax}(W^{\top}\boldsymbol{z_i} + b)[\boldsymbol{y}]
        \text{ where } \boldsymbol{y}\in\boldsymbol{\mathcal{Y}}.
    \end{gathered}
\end{equation*}
Using $p(\boldsymbol{y}|\boldsymbol{z}; \phi)$, we can estimate $U_t$ and $U_p$, and then $U_e$:
\begin{equation}
    \begin{split}
        U_e \approx \underbrace{H\big[\frac{1}{T} \sum\limits_{t=1}^T p(\boldsymbol{y}|\boldsymbol{z_t}; \phi)\big]}_{U_t(\boldsymbol{y}|\boldsymbol{x}, \boldsymbol{\mathcal{D}})} - 
        \underbrace{\frac{1}{T} \sum\limits_{t=1}^T H \big[p(\boldsymbol{y}|\boldsymbol{z_t}; \phi)\big]}_{U_p(\boldsymbol{y}|\boldsymbol{x}, \theta)},
    \end{split}
    \label{eq:mi_appox}
\end{equation}
where $\boldsymbol{z_t} \sim \mathcal{N}(h, \text{diag}(\sigma\odot\sigma)), h = g_{\bar{\theta}}(\boldsymbol{x})$ and $\sigma = \mathcal{C}_{\phi}(h)$.

\noindent To motivate the intuition behind $\mathcal{C}_{\phi}$, we use $g_{\bar{\theta}}$ to embed $\boldsymbol{x}$ into the low-dimensional predictive mean $h$, which we then feed to $\mathcal{C}_{\phi}$ to get the predictive standard error $\sigma$. Here, $g_{\bar{\theta}}$ learns the predictive features of $\boldsymbol{x}$ while $\mathcal{C}_{\phi}$ captures the uncertainty associated with their predictiveness. As such, we can use $h$ and $\sigma$ to parameterize a normal distribution and get amortized estimates of the predictive uncertainty.

\subsection{Generating Adversarial Examples}
To generate adversarial examples using our uncertainty threat model, we optimize the objective $\boldsymbol{\mathcal{L}}$ defined below:
\begin{equation}
    \begin{split}
         \boldsymbol{\mathcal{L}}(\boldsymbol{x}, \boldsymbol{y}) \triangleq& \max_{\boldsymbol{\omega}\sim p_{\nu}(\boldsymbol{x})}\bigg(U_e(\star, \boldsymbol{y}|\boldsymbol{\mathcal{D}}, \boldsymbol{\omega}\odot \boldsymbol{x})\\
         +& \max\limits_{\boldsymbol{x'}\in\mathcal{S}(f(\boldsymbol{x},\, \boldsymbol{\omega}))}\boldsymbol{\mathcal{L}_{\text{adv}}}(\boldsymbol{x'}, \boldsymbol{y}) - \lambda\cdot\|\boldsymbol{\omega}\|_1\bigg).
    \end{split}
    \label{eq:obj}
\end{equation}
$U_e(\star, \boldsymbol{y}|\boldsymbol{\mathcal{D}}, \boldsymbol{\omega}\odot \boldsymbol{x})$ is the epistemic uncertainty on $\boldsymbol{\omega}\odot \boldsymbol{x}$ where $\star$ denotes either $\theta$ if the classifier $g_{\theta}$ is stochastic, or $\phi$ if $g_{\theta}$ is deterministic. To compute $U_e(\star, \boldsymbol{y}|\boldsymbol{\mathcal{D}}, \boldsymbol{\omega}\odot \boldsymbol{x})$, we can replace $\boldsymbol{x}$ by $\boldsymbol{\omega}\odot \boldsymbol{x}$ either in Equation~\ref{eq:mi2} or Equation~\ref{eq:mi_appox}. $\boldsymbol{\mathcal{L}_\text{adv}}$ is defined in Equation~\ref{eq:unc}. Here, $p_{\nu}$ denotes the distribution over all possible masks $\boldsymbol{\omega}$. $\mathcal{S}(f(\boldsymbol{x},\, \boldsymbol{\omega}))$, defined in Equation~\ref{eq:unc}, denotes the set of permissible perturbations we can apply to $\boldsymbol{x}$ in order to induce misclassification from $g_{\theta}$.
 \vskip 0.1cm

\noindent To learn the distribution $p_{\nu}$, we train a feedforward network called mask model. This network takes as input an image $\boldsymbol{x}$ and outputs probabilities for each point in $\boldsymbol{x}$ (across all its channels). We use the probabilities to parameterize a Bernoulli distribution from which we then sample binary masks. A mask $\boldsymbol{\omega}\sim p_{\nu}(\boldsymbol{x})$ is of the same size as $\boldsymbol{x}$ with a one-to-one coordinate mapping with $\boldsymbol{x}$. We regularize the mask model so that $\boldsymbol{\omega}$ is neither full, that is; $\omega_1=...=\omega_n=1$, nor completely sparse, i.e.; $\omega_1=...=\omega_n=0$. To that end, we add a sparsity penalty $\|\boldsymbol{\omega}\|_1$ weighted by $\lambda$ to control its effect on the remaining part of the objective $\boldsymbol{\mathcal{L}}$. 
\begin{table*}[t]
\caption{Accuracy of defended classifiers before and after \textit{delta} and ReColor attacks, under their natural threat model and ours.}
\label{tab:asr}
\vskip -0.5in
\begin{center}
\begin{small}
\begin{sc}
\begin{tabular}{
    >{\arraybackslash}m{0.3cm}|
    >{\arraybackslash}m{3.0cm}|
    >{\centering\arraybackslash}m{1.9cm}
    >{\centering\arraybackslash}m{1.9cm}
    >{\centering\arraybackslash}m{1.9cm}|
    >{\centering\arraybackslash}m{1.9cm}
    >{\centering\arraybackslash}m{1.9cm}}
\toprule
&\multirow{2}{*}{Attacks $\downarrow$}         & \multicolumn{3}{c|}{\textit{Deterministic Defenses} (\%)} & \multicolumn{2}{c}{\textit{Bayesian Defenses} (\%)}\\
\cmidrule{3-7}
& &None & \textit{delta} & \textit{StdAdv}  & None & \textit{delta} \\
\midrule
\parbox[t]{10mm}{\multirow{2}{*}{\rotatebox[origin=c]{90}{\textit{MNIST}}}} 
&Vanilla               & 98.89              & 98.64            & 97.33             & 97.91             & 76.76\\
&\textit{delta}        & \;\;\textbf{0.00}  & \textbf{92.47}   & \;\;\textbf{0.00} & \;\;\textbf{0.00} & \;\;\textbf{0.00}\\
&Unc. + \textit{delta} & \;\;\textbf{0.00}  & 94.79            & \;\;\textbf{0.00} & \;\;\textbf{0.00} & \;\;\textbf{0.00}\\
\midrule
\parbox[t]{10mm}{\multirow{5}{*}{\rotatebox[origin=c]{90}{\textit{CIFAR-10}}}} 
&Vanilla               & 92.20              & 84.80            & 82.88 & 89.62            & 77.96 \\
&\textit{delta}        & \;\;\textbf{0.00}  & \textbf{30.60}   & \;\;\textbf{0.00} & \;\;\textbf{0.00}& \;\;\textbf{1.38}\\
&Unc. + \textit{delta} & \;\;\textbf{0.00}  &  33.32   & \;\;\textbf{0.00} & \;\;0.13& \;\;3.26\\ 
&ReColor               & \textbf{63.76}     & \textbf{75.12}           & \textbf{68.62}    & \textbf{83.40}         & \textbf{69.58} \\
&Unc. + ReColor        &  65.69             & 78.64  & \textbf{68.62}  &  85.19            & 71.18  \\
\midrule
\parbox[t]{10mm}{\multirow{5}{*}{\rotatebox[origin=c]{90}{\textit{STL-10}}}} 
&Vanilla               & 81.37              & 63.42            & \NA   & 74.85             & 59.89 \\
&\textit{delta}        & \;\;\textbf{0.00}  & \,\,\textbf{0.00}& \NA   & \;\;\textbf{0.00} & \;\textbf{0.66}\\
&Unc. + \textit{delta} & \;\,\textbf{0.00}  & \,\,\textbf{0.00}& \NA   & \;\;\textbf{0.00} & \;\;1.44\\ 
&ReColor               & \textbf{53.51}     & \textbf{44.72}   & \NA   & \textbf{15.25}    & \textbf{18.55} \\
&Unc. + ReColor        & 57.23              & 45.26            & \NA   & 17.07             & 21.03\\ 
\bottomrule
\end{tabular}
\end{sc}
\end{small}
\end{center}
\vskip -0.25in
\end{table*}
\begin{table*}[t]
\caption{\small{Mask Sparsity ($\ell_0$). For vanilla \textit{delta} and ReColor, we get the score by computing the $\ell_0$ norm between an input and its adversarial example divided by $C\times W\times H$ where $C$ is the number of channels, $W$ the width and $H$ the height of the images. \textbf{Lower is better.}}}
\label{tab:sparsity}
\begin{center}
\begin{small}
\begin{sc}
\begin{tabular}{
    >{\arraybackslash}m{0.3cm}|
    >{\arraybackslash}m{3.0cm}|
    >{\centering\arraybackslash}m{1.9cm}
    >{\centering\arraybackslash}m{1.9cm}
    >{\centering\arraybackslash}m{1.9cm}|
    >{\centering\arraybackslash}m{1.9cm}
    >{\centering\arraybackslash}m{1.9cm}}
\toprule
&\multirow{2}{*}{Attacks $\downarrow$}         & \multicolumn{3}{c|}{\textit{Deterministic Defenses} (\%)} & \multicolumn{2}{c}{\textit{Bayesian Defenses} (\%)}\\
\cmidrule{3-7}
&                      &None      & \textit{delta} & \textit{StdAdv}  & None & \textit{delta} \\
\midrule
\parbox[t]{10mm}{\multirow{2}{*}{\rotatebox[origin=c]{90}{\textit{MNIST}}}} 
&\textit{delta}        & 88.61         & 98.73         & 93.11         & 88.83          & 89.20         \\[1ex]
&Unc. + \textit{delta} & \textbf{74.92}& \textbf{79.88}& \textbf{73.11}           & \textbf{69.82} & \textbf{75.98}\\[1ex]
\midrule
\parbox[t]{10mm}{\multirow{4}{*}{\rotatebox[origin=c]{90}{\textit{CIFAR-10}}}} 
&\textit{delta}        & 86.92         & 78.61           & 89.33         & 99.24          & 99.70 \\
&Unc. + \textit{delta} & \textbf{55.40}& \textbf{55.68}  & \textbf{86.97}& \textbf{62.66} & \textbf{62.65} \\ 
&ReColor               & 99.55         & 100.0           & 100.0         & 100.0          & 100.0 \\
&Unc. + ReColor        & \textbf{55.75}& \textbf{55.78}  & \textbf{89.21}& \textbf{46.42} & \textbf{46.35} \\
\midrule
\parbox[t]{10mm}{\multirow{4}{*}{\rotatebox[origin=c]{90}{\textit{STL-10}}}} 
&\textit{delta}        & 43.38          & 75.82     & \NA           & 99.43          & 100.0 \\
&Unc. + \textit{delta} & \textbf{28.14}             & \textbf{27.17}& \NA            & \textbf{43.89} & \textbf{58.93} \\ 
&ReColor               & 99.23          & 100.0     & \NA           & 100.0          & 100.0 \\
&Unc. + ReColor        & \textbf{78.12} & \textbf{58.40}            & \NA           & \textbf{58.93} & \textbf{58.93} \\ 
\bottomrule
\end{tabular}
\end{sc}
\end{small}
\end{center}
\vskip -0.2in
\end{table*}
\section{Experiments \& Results}
We wish to show that under our threat model both $\ell_p$ and functional attacks can retain similar attack strengths while perturbing the inputs less than under their own threat model. In that regard, we seek to answer the following questions through our experiments: (Q1) \textit{How strong are the attacks?} (Q2) \textit{How imperceivable are the adversarial examples?} (Q3) \textit{How transferable are the attacks?}  Thus, we validate the attacks under our threat model based on three criteria: \textit{attack success rate, imperceptibility, and transferability.}
\vskip 0.1cm

\noindent \textbf{Datasets.} Taking inspiration from similar studies, we conduct experiments on CIFAR-10~\cite{CIFAR10}, MNIST~\cite{lecun-mnisthandwrittendigit-2010}, and STL-10~\cite{coates11a}. For a detailed overview of the experimental setup and more results, we refer the reader to Appendix~\ref{sec:experiments} and~\ref{sec:results}.
 \vskip 0.1cm

\noindent \textbf{Attack Success Rate.} We define \textit{attack success rate} as the decrease in test accuracy of a classifier resulting from the misclassifications that our perturbations induce. Here, we ask ourselves the following: \textit{Can delta or ReColor retain their attack success rates under our threat model while perturbing the inputs much less?} To answer this question, we evaluate the sparsity of the binary masks we generate using the $\ell_0$ \textit{norm as a sparsity measure}.
 \vskip 0.1cm

\noindent We report in Table~\ref{tab:asr} the attack success rates of \textit{delta} and ReColor attacks under their threat model and under ours. The results in Table~\ref{tab:asr} are to be read side by side with the sparsity scores in Table~\ref{tab:sparsity}. \textsc{Vanilla} denotes the targeted classifier with \textsc{None} reflecting its accuracy when it is undefended. \textit{delta} and \textit{StdAdv} show the drop in accuracy after attacking \textsc{Vanilla} when defended either with \textit{delta} or \textit{StdAdv}. As the results show, \textit{delta} and ReColor achieve similar attack success rates under our threat model while perturbing the inputs consistently less than under their own. We explore in Appendix the effects of the mask models on the attacks.
\vskip 0.1cm

\noindent Both \textit{delta} and ReColor achieve better imperceptibility under our threat model than under their native ones, as evidenced by our results in Table~\ref{tab:lpips} in Appendix~\ref{sec:results}. In Table~\ref{tab:transferability} in Appendix~\ref{sec:results}, we also show that the reduced number of perturbed pixels do not seem to significantly impact the transferability of the generated adversarial examples.

\section{Conclusion}
\noindent We presented \textit{uncertainty attacks}, a novel class of adversarial attacks against deep learning models. Our attack model utilizes the uncertainty associated with the predictions of a stochastic classifier or the amortized uncertainty of its surrogate to localize regions in the inputs to perturb so as to induce misclassification. Unlike standard attacks which perturb inputs indiscriminately, we focus only on carefully chosen regions, yielding thus improved imperceptibility. Mainstream attacks can still be evaluated under our threat model, both against defended and undefended classifiers, and produce strong adversarial examples that retain actually a greater degree of similarity with the original inputs.

{\small
\bibliographystyle{ieee_fullname}
\bibliography{egbib}
}

\onecolumn
\appendix

\section{Motivations}
\label{sec:motivations}
\vskip 0.1in
\noindent Deep learning models continue to achieve impressive performances on a variety of challenging machine learning tasks~\cite{alphafold,brown2020language}. However, they remain fragile. Small, seemingly inconspicuous noise injected to the input data can cause a deep learning model to suddenly make erroneous predictions with high confidence~\cite{ianj.goodfellow2014,42503}. As deep learning models permeate different and vital segments of the technology industry, such susceptibility to adversarial perturbations is a cause for concern and calls upon machine learning practitioners to revisit their evaluation protocol.

\vskip 0.1in
\noindent Recently, many methods have been proposed to scrutinize deep learning models and expose their weaknesses in areas as diverse as computer vision~\cite{DBLP:journals/corr/abs-1802-08195,DBLP:journals/corr/abs-1712-09665}, speech recognition~\cite{DBLP:journals/corr/abs-1808-05665,cisse2017houdini}, and machine translation~\cite{ebrahimi-etal-2018-adversarial}.\footnote{For a detailed review of the adversarial methods, we refer the interested reader to~\cite{DBLP:journals/corr/abs-1911-05268} and \cite{xu2019adversarial}.} These methods known as adversarial attacks aim generally at producing adversarial examples that can mislead a model while being perceptually similar to the original inputs~\cite{DBLP:journals/corr/abs-1906-00001,2019arXiv190303905A,DBLP:journals/corr/abs-1907-02044,madry2019deep,2019arXiv190208265J,DBLP:journals/corr/abs-1802-09653,athalye2017,42503}.

\vskip 0.1in
\noindent By and large, most of the adversarial attacks~\cite{DBLP:journals/corr/abs-1802-00420,DBLP:journals/corr/abs-1712-02779,DBLP:journals/corr/CarliniW16a} and defense methods~\cite{DBLP:journals/corr/abs-1907-10764, DBLP:journals/corr/abs-1903-06603,sinha2018certifiable, DBLP:journals/corr/abs-1801-09344,DBLP:journals/corr/abs-1711-00851,ianj.goodfellow2014} consider \textit{additive $\ell_p$ threat models} where adversarially crafted examples and their natural inputs differ by a small $\ell_p$ distance. Concretely, the degree to which an adversarial example is perceptually similar to its original input is measured in terms of an $\ell_p$ distance as a substitute proxy of human perception of similarity. 

\begin{figure}[th]
    \vskip -0.2in
    \centering
    \subfloat{{\includegraphics[width=4cm]{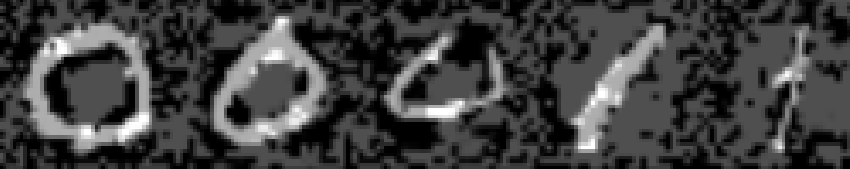}}}
    \qquad
    \subfloat{{\includegraphics[width=4cm]{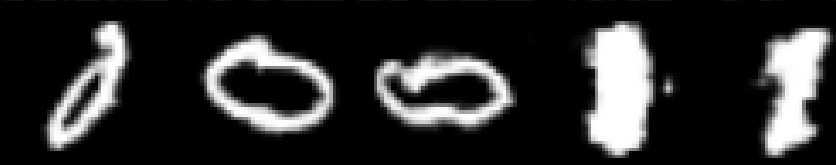}}}
	\caption{These adversarial images are supposed to represent the digits 0 and 1, yet they look heavily distorted, blurry, and not quite legitimate 0s or 1s. The images are taken from \cite{song2018constructing}.}
	\label{fig:samples1}
    \vskip -.15in
\end{figure}

\begin{figure}[ht]
    \centering
    \includegraphics[width=7.5cm]{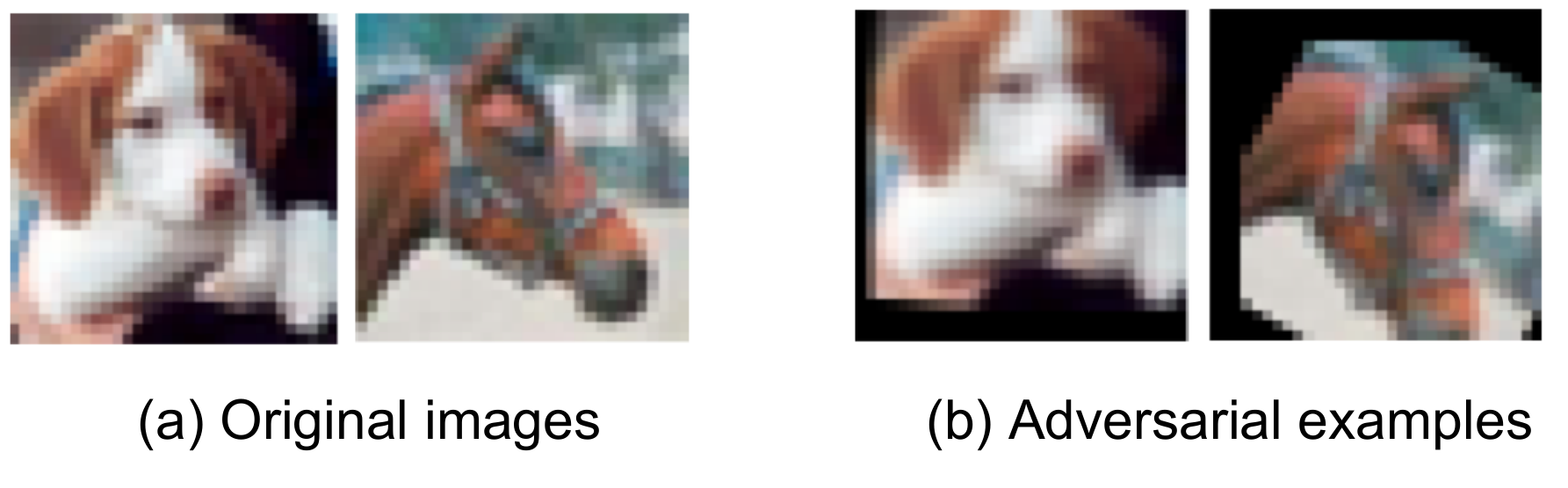}
    \vspace{-0.1in}
    \caption{Adversarial examples created using geometric distortions may look perceptually to the inputs but they have large $\ell_p$ distances in the input space. Images taken from~\cite{DBLP:journals/corr/abs-1712-02779}.}
    \label{fig:samples2}
    \vskip -0.15in
\end{figure}

\vskip 0.1in
\noindent However, it is well documented that \textit{nearness} according to an $\ell_p$ norm is unsuitable for measuring perceptual similarity when creating adversarial examples~\cite{DBLP:journals/corr/abs-1802-09653}. In the image domain specifically, clipping, rotation, occlusions, changes of color or illumination in an image, to name a few, do not always preserve the true nature of the image~\cite{DBLP:journals/corr/abs-1802-09653}. \textit{Oftentimes, the resulting images appear perceptually distorted, blurry, or unrealistic} as exemplified in Figure~\ref{fig:samples1}. In Figure~\ref{fig:samples2}, we can also notice two adversarial images that appear similar to their original inputs, yet their $\ell_p$ distances are --- $\ell_0\geq 3,10$, $\ell_2\geq 15.83$, and $\ell_\infty\geq 0.87$~\cite{DBLP:journals/corr/abs-1802-09653} --- much higher than the $\ell_p$ norm thresholds used in most of the existing attacks.

\vskip 0.1in
\noindent Although quantifying imperceptibility can be challenging due to the subjective nature of human perception, just using $\ell_p$ norms as proxies may hinder the construction of adversarial examples that preserve the key features of the inputs. Such an outcome is not only desirable for probing deep learning models in order to detect their weaknesses, it is equally important for increasing their robustness to such attacks. Indeed, many defenses underlied by the $\ell_p$ threat models tend to be brittle~\cite{DBLP:journals/corr/abs-1802-00420}. When they appear robust, their robustness does not always generalize to larger yet similar $\ell_p$ perturbations, nor does it extrapolate to different threat models~\cite{2019arXiv191006259S}. A failure to investigate more inconspicuous attacks could cause major impediments in deploying and entrusting deep learning models with making decisions in security-sensitive applications.

\vskip 0.1in
\noindent In this paper, we present \textit{localized uncertainty attacks}, a novel class of threat models for creating adversarial examples against deep learning models. Under this threat model, adversarial examples are generated via \textit{replacement} by localizing regions in the original inputs where a classifier is \textit{uncertain or makes unconstrained extrapolations}. By ``replacement'', we mean that instead of perturbing indiscriminately an input as standard threat models intend to, \textit{only select regions or points} in the input are subject to adversarial changes. Our attack model can essentially be decomposed into: \textit{region proposals} and \textit{adversarial perturbations}.

\begin{figure}[t!]
\vskip -0.06in
\begin{center}
    \includegraphics[width=7cm]{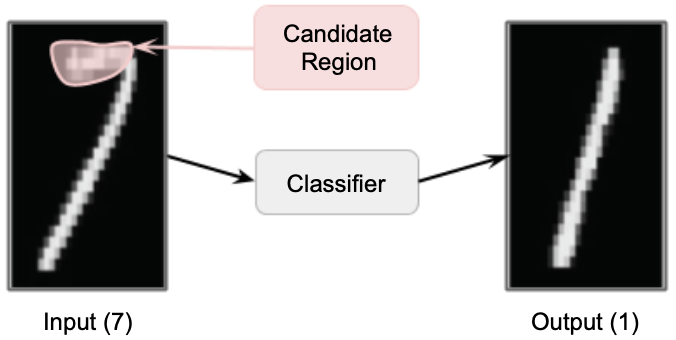}
\end{center}
\caption{On the left we have a digit 7 to which is overlaid a region proposal colored in pink. Grayscaling this pink region can induce misclassification as the classifier may interpret the input as digit 1.}
\label{fig:proposal}
\vskip -.2in
\end{figure}

\vskip 0.1in
\noindent \textbf{Region Proposals.} Given an input, we learn which regions in the input to perturb. \textit{A region is a set of points which may or may not cluster together}. For instance, a region could denote a patch or a set of non-adjacent pixels in an image, filterbank energies in an audio sample, or a set of words in a sentence. We prioritize mainly the regions where the target classifier is poorly constrained. For illustration, we give a high-level depiction of such a region in Figure~\ref{fig:proposal}. To select which regions to perturb, we learn a model which when given the input produces a \textit{binary mask} of similar shape as the input. The mask maintains a one-to-one coordinate mapping with the input. Explicitly, every element of the mask is mapped to the same coordinate point in the input and dictates whether to perturb that point or to leave it unaltered.

\noindent \textbf{Adversarial Perturbations.} Our uncertainty threat model can be applied to a wide variety of spatio-temporal modalities including images, text, combination thereof (e.g., OCR), or speech. In this paper, however, we focus only on images. Our threat model extends additive and functional threat models. For instance, we can leverage the losses in \cite{DBLP:journals/corr/CarliniW16a} to perturb input images. Similar to~\cite{DBLP:journals/corr/abs-1906-00001}, we can perturb pixels using parameterized functions to transform their color space uniformly in order to induce misclassification. Unlike these threat models, however, we do not perturb arbitrarily every pixel in an image. \textit{We perturb only the select few pixels that our masking algorithm recommends. As a result, we achieve improved imperceptibility over additive and functional attacks}.

\vskip 0.1in
\noindent For evaluation, we consider additive and functional attacks under our threat model and target undefended and defended classifiers --- which are either deterministic or stochastic. We find that under our threat model both attacks are strong attacks, lowering the accuracy of a ResNet-32 trained on CIFAR-10 to 0.0\% while perturbing 31\% less than under a typical $\ell_p$ threat model. The combination reduces in one instance the accuracy of an STL-10 classifier to 0.0\% while perturbing 43\% less than under a standard $\ell_p$ threat model.

\vskip 0.1in
\noindent \textbf{Contributions.} To summarize our contributions, we propose localized uncertainty attacks, a novel class of threat models for creating adversarial examples in the image domain. Unlike standard threat models, which perturb indiscriminately every input, our attack model focuses only on select few pixels; resulting thus in improved imperceptibility while maintaining similar attack success rates. Our perturbation hinges on quantifying the uncertainty associated with the predictions of the target classifiers on specific regions of the inputs. For deterministic classifiers, we train surrogate uncertainty models to get amortized uncertainty estimates. To the authors' knowledge, there is no or limited work on using uncertainty to attack both \textit{deterministic and stochastic} deep learning models. Our threat model can also be coupled with standard threat models to achieve even stronger attacks.

\section{Related Work}
\noindent In this section, we review some of the current studies that are most closely related to our approach. For a more detailed review of adversarial attacks, we refer the interested reader to~\cite{DBLP:journals/corr/abs-1911-05268,xu2019adversarial,8611298}.

\vskip 0.1in
\noindent \textbf{Adversarial Examples} have received in recent years renewed interests resulting in a number of approaches for generating novel adversarial attacks and defending against them. Many adversarial attacks focus on finding small, inconspicuous perturbations that can deceive deep learning models while being imperceptible to humans~\cite{wong2020learning,2019arXiv190208265J,DBLP:journals/corr/abs-1802-08195}. However, several other notions of adversarial perturbations have been studied~\cite{bhattad2020unrestricted,DBLP:journals/corr/abs-1712-02779,song2018constructing}. Our approach falls in the category of attacks that look to craft adversarial examples with minimal distortion to the inputs. In this study, we proposed a new method to improve the imperceptibility of adversarial perturbations by carefully selecting which regions in an input to perturb. 

\vskip 0.1in
\noindent \textbf{Uncertainty Estimation.} To select which regions in an input to perturb, we localize the regions where a classifier appears uncertain or poorly constrained. Taking inspiration from~\cite{smith2018understanding}, we use entropy in the probability space as a proxy for classification uncertainty. Using this proxy, we can retrieve the epistemic uncertainty of a stochastic classifier and perform gradient-based attacks. We extend this approach of quantifying epistemic uncertainty to deterministic classifiers using surrogate models to get amortized uncertainty estimates with minimal computational overhead.

\vskip 0.1in
\noindent \cite{carbone2020robustness} have argued that Bayesian Neural Networks are robust to gradient-based attacks. However, the conditions under which their findings hold are hard to observe in practice. To the author's knowledge, there is no or limited study that uses uncertainty to perform gradient-based attacks against both deterministic and stochastic deep learning models. Owing to its simplicity, our attack model can be easily combined with a variety of attack models including \textit{delta}~\cite{madry2019deep,DBLP:journals/corr/CarliniW16a}, functional~\cite{DBLP:journals/corr/abs-1906-00001}, and spatial transformation attacks~\cite{DBLP:journals/corr/abs-1801-02612,DBLP:journals/corr/abs-1712-02779} to minimize perceptual distortions without sacrificing strength. 

\section{Experimental Setup}
\label{sec:experiments}
\noindent \textbf{Experimental Setup.} To validate our uncertainty threat model, we consider deterministic and stochastic classifiers. 

\vskip 0.1in
\noindent \textit{Deterministic Classifiers.} For CIFAR-10, we consider an undefended ResNet-32~\cite{DBLP:journals/corr/HeZRS15}, a ResNet-32 robustified with \textit{delta} attacks, and a ResNet-32 defended with StdAdv~\cite{DBLP:journals/corr/abs-1801-02612}. We use the classifiers pretrained by~\cite{DBLP:journals/corr/abs-1906-00001}. For MNIST, we evaluate with an undefended LeNet5~\cite{Lecun98gradient-basedlearning}, a LeNet5 trained with StdAdv, and a LeNet5 trained with PGD~\cite{madry2019deep}. For STL-10, we experiment with the undefended and defended classifiers pretrained by~\cite{DBLP:journals/corr/abs-1801-02612}.

\vskip 0.1in
\noindent For each of the deterministic classifiers, we train a surrogate uncertainty model to get amortized uncertainty estimates which we then use to train a separate mask model. Note that the surrogate model and the mask model are trained jointly. The surrogate model is a feedforward network trained with dropout. The mask model is an hourglass model consisting of convolutional layers. We use a reparameterized Tanh over the outputs of the mask model to define a (relaxed) Bernoulli distribution from which we sample the binary masks $\boldsymbol{\omega}$.

\vskip 0.1in
\noindent \textit{Stochastic Classifiers.} We experiment with MNIST, CIFAR-10 and STL-10 where we evaluate our attack model against Bayesian classifiers. For MNIST, we experiment with an undefended Bayesian LeNet5 that we train using mean-field variational inference~\cite{blundell2015weight}, and one defended with PGD. For CIFAR-10, we consider two Bayesian VGGs~\cite{Simonyan15}: an undefended VGG and one defended with \textsc{Adv-BNN}~\cite{DBLP:journals/corr/abs-1801-02612} both pretrained by ~\cite{DBLP:journals/corr/abs-1801-02612}. For STL-10, we experiment with the models provided by~\cite{DBLP:journals/corr/abs-1801-02612}.

\vskip 0.1in
\noindent \textbf{Perturbations.} Under our threat model, we evaluate the \textit{delta} attacks which use an $\ell_\infty$ additive threat model with $\epsilon=8/255$. Under also our threat model, we evaluate the functional attacks~\cite{DBLP:journals/corr/abs-1906-00001} which transform uniformly the color space of all the pixels of an image. In what follows, \textit{delta} refers to the additive perturbations, and ReColor to the trilinear transformation of~\cite{DBLP:journals/corr/abs-1906-00001} in the RGB color space where we relax their smoothness constraint. For the latter, we allow each channel of a color to change by $8/255$ at most, that is; $\epsilon_R=\epsilon_G=\epsilon_B=0.03$. We use Projected Gradient Descent or PGD~\cite{madry2019deep} with 100 iterations to carry out the uncertainty attacks. Finally, we consider as adversarial loss the $f_6$ loss from~\cite{DBLP:journals/corr/CarliniW16a} with $\kappa=10$. 

\section{More Evaluation Results}
\label{sec:results}
\subsection{Surrogate Models}
\begin{figure}[!ht]
\vskip -0.1in
\begin{center}
    \includegraphics[width=7.33cm]{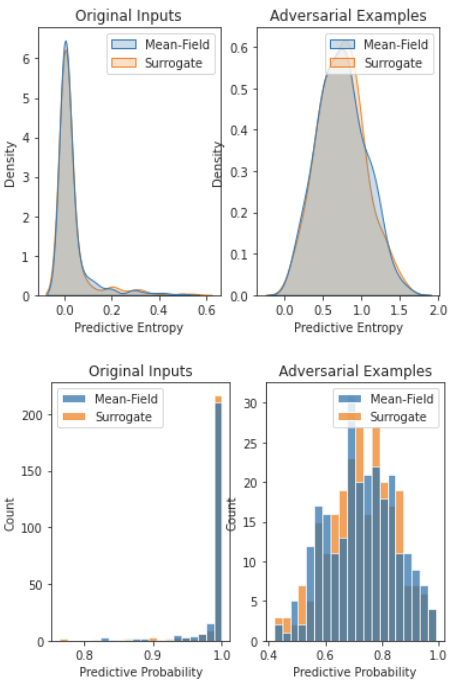}
\end{center}
\caption{Mean-Field model compared to a LeNet surrogate.}
\label{fig:xentr}
\vskip -0.1in
\end{figure}

\noindent To attack deterministic classifiers, we train surrogate models to get amortized uncertainty estimates. A surrogate model is two-layers deep and is trained jointly with the mask model to minimize overhead. To validate empirically these models, we compare in Figure~\ref{fig:xentr} a Mean-Field with the surrogate of a deterministic LeNet5 based on their predictive probability and entropy on hold-out MNIST images and their adversaries. As Figure~\ref{fig:xentr} shows, the surrogate approximates well the Mean-Field on both the inputs and their adversaries. 

\begin{figure*}[ht]
\vskip -0.1in
\begin{center}
    \subfloat{{\includegraphics[width=8.1cm]{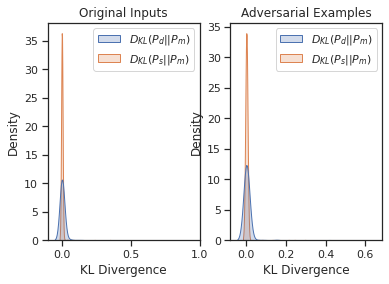} }}%
    \qquad
    \subfloat{{\includegraphics[width=8.1cm]{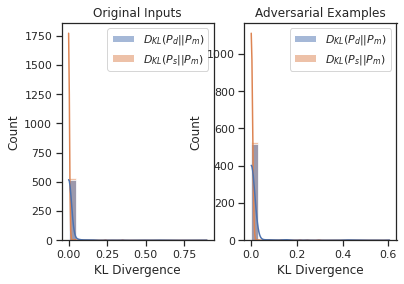} }}%
    \qquad
    \qquad
\begin{small}
\begin{sc}
\begin{tabular}{
    >{\centering\arraybackslash}m{3.2cm}
    >{\centering\arraybackslash}m{3.2cm}|
    >{\centering\arraybackslash}m{3.2cm}
    >{\centering\arraybackslash}m{3.2cm}}
\toprule
\multicolumn{2}{c|}{$D_{KL}(P_d\|P_m)$} & \multicolumn{2}{c}{$D_{KL}(P_s\|P_m)$}\\
\midrule
\textit{Natural Inputs} & \textit{Adversarial Examples} & \textit{Natural Inputs} & \textit{Adversarial Examples} \\
\midrule 
5.14                    & 4.44                          & \textbf{0.82}           & \;\;\textbf{1.07}   \\
\bottomrule
\end{tabular}
\end{sc}
\end{small}
    \caption{We compare the surrogate model with the Mean-field using KL divergence. As the plots show, the surrogate model }%
    \label{fig:density_plots}.
\end{center}
\end{figure*}

\vskip 0.1in
\noindent We extend this empirical validation further using KL divergence conditioned on the predictive distribution of the deterministic LeNet5 to measure the discrepancy between both models based on their distributions. Formally, if we denote the predictive distributions of the deterministic LeNet5, the surrogate model, and the Mean-field respectively by $P_d$, $P_s$, and $P_m$, we compare the surrogate and the Mean-field based on the divergences $D_{KL}(P_d\|P_m)$ and $D_{KL}(P_s\|P_m)$ both on the natural MNIST images and their adversarial examples. We show in Figure~\ref{fig:density_plots} the kernel density estimation plots for both divergences and summarize the aggregated KL in the table nested in Figure~\ref{fig:density_plots}. As we can notice, here too the surrogate model approximates quite well the Mean-field model as $D_{KL}(P_s\|P_m)$ is much smaller than $D_{KL}(P_d\|P_m)$.

\vskip 0.1in
\noindent We show in Algorithm~\ref{alg:attacks} how we train the surrogate model along with the mask model when generating adversarial examples. In all our experiments, we work with two-layers deep feedforward neural networks as surrogate models. As stated in the main paper, the mask model is an hourglass model consisting of convolutional layers. We use a reparameterized Tanh over the outputs of the mask model to define a Bernoulli distribution from which we sample the masks. As the masks consist of binary values, in order to perform backpropagation onto the mask model, we relax the Bernoulli distribution using the Gumbel-Max trick.
\begin{algorithm}[H]
    \begin{algorithmic}
      \State{{\bfseries Require:} Target classifier $g_{\theta}$}
      \State{{\bfseries Require:} Learning rates $\alpha$ and $\beta$}
      \For{$(\boldsymbol{x}, \boldsymbol{y})\sim\boldsymbol{\mathcal{D}}$}
        \State{ // \textit{This for loop applies only when $g_{\theta}$ is deterministic.}}
        \State{Draw randomly a mini-batch $\boldsymbol{\mathcal{B}}$ of examples from $\boldsymbol{\mathcal{D}}$}
        \For{$(\boldsymbol{x_c}, \boldsymbol{y_c})\sim\boldsymbol{\mathcal{B}}$}
            \State{$h_c\leftarrow g_{\bar{\theta}}(\boldsymbol{x_c})$;\, $\sigma\leftarrow \mathcal{C}_{\phi}(h_c)$}
            \State{Sample $\boldsymbol{z}_1, ..., \boldsymbol{z}_N\sim \mathcal{N}(h_c, \text{diag}(\sigma\odot\sigma))$}
            \State{$\phi\leftarrow \phi + \alpha\cdot\nabla_{\phi} \sum_{i=1}^N \log p(\boldsymbol{y_c}|\boldsymbol{z_i}, \phi)$}
        \EndFor
      \State{Sample a binary mask $\boldsymbol{\omega}\sim p_{\nu}(\boldsymbol{x})$}
      \State{$\boldsymbol{x'}\leftarrow \argmax_{\boldsymbol{x'}\in\mathcal{S}(f(\boldsymbol{x},\, \boldsymbol{\omega}))} \ell(g_\theta(\boldsymbol{x'}), \boldsymbol{y})$}
      \State{Perform a gradient update on $\nu$ using Equation 7 (see main paper)}
      \EndFor
\end{algorithmic}
  \caption{Generating Adversarial Examples.}
  \label{alg:attacks}
\end{algorithm}
\subsection{Mask Ablation Study}
\noindent To evaluate the efficacy of our masking mechanism, we do an ablation study. We replace the mask models trained with uncertainty with a random distribution to parameterize a Bernoulli distribution from which we sample the binary masks. We conduct experiments using this ablation against MNIST and CIFAR-10 classifiers and report the results in Table~\ref{tab:ablation}. As the results indicate, both \textit{delta} and ReColor perform worse under this setting than under our uncertainty threat model. 
\begin{table*}[ht]
\caption{Model accuracies before and after \textit{delta} and ReColor attacks, under their native threat model and ours compared with the baseline.}
\label{tab:ablation}
\vskip -0.2in
\begin{center}
\begin{small}
\begin{sc}
\begin{tabular}{
    >{\arraybackslash}m{0.3cm}|
    >{\arraybackslash}m{3.0cm}|
    >{\centering\arraybackslash}m{1.9cm}
    >{\centering\arraybackslash}m{1.9cm}
    >{\centering\arraybackslash}m{1.9cm}|
    >{\centering\arraybackslash}m{1.9cm}
    >{\centering\arraybackslash}m{1.9cm}}
\toprule
&\multirow{2}{*}{Attacks $\downarrow$}         & \multicolumn{3}{c|}{\textit{Deterministic Defenses} (\%)} & \multicolumn{2}{c}{\textit{Bayesian Defenses} (\%)}\\
\cmidrule{3-7}
& &None & \textit{delta} & \textit{StdAdv}  & None & \textit{delta} \\
\midrule
\parbox[t]{10mm}{\multirow{4}{*}{\rotatebox[origin=c]{90}{\textit{MNIST}}}} 
&Vanilla               & 98.89              & 98.64            & 97.33             & 97.91             & 76.76\\
\cmidrule{2-7}
&\textit{delta}        & \;\;\textbf{0.00}  & \textbf{92.47}   & \;\;\textbf{0.00} & \;\;\textbf{0.00} & \;\;\textbf{0.00}\\
&Unc. + \textit{delta} & \;\;\textbf{0.00}  & 94.79            & \;\;\textbf{0.00} & \;\;\textbf{0.00} & \;\;\textbf{0.00}\\
&Rand. + \textit{delta}& 41.16 & 97.88          & \;\;6.86            & 25.87& 10.71\\ 
\midrule
\parbox[t]{10mm}{\multirow{8}{*}{\rotatebox[origin=c]{90}{\textit{CIFAR-10}}}} 
&Vanilla               & 92.20             & 84.80          & 82.88 & 89.62     & 77.96 \\
\cmidrule{2-7}
&\textit{delta}        & \;\;\textbf{0.00} & \textbf{30.60} & \;\;\textbf{0.00} & \;\;\textbf{0.00}& \;\;\textbf{1.38}\\
&Unc. + \textit{delta} & \;\;\textbf{0.00} & 33.32          & \;\;\textbf{0.00}& \;\;0.13& \;\;3.26\\ 
&Rand. + \textit{delta}& \;\;\textbf{0.00} & 61.53          & 76.51            & \;\;4.92& 29.25\\ 
\cmidrule{2-7}
&ReColor               & \textbf{63.76}    & \textbf{75.12} & \textbf{68.62}   & \textbf{83.40}  & \textbf{69.58} \\
&Unc. + ReColor        & 65.69             & 78.64          & \textbf{68.62}   &  85.19          & 71.18  \\
&Rand. + ReColor       & 83.13             & 83.18          & 75.86            &  86.57& 77.65\\ 
\bottomrule
\end{tabular}
\end{sc}
\end{small}
\end{center}
\vskip -0.15in
\end{table*}

\subsection{Imperceptibility}
\noindent We evaluate how visually imperceivable the adversarial examples that \textit{delta} and ReColor generate, under our threat model and under their own, based on: \textit{i}.) the sparsity of the binary masks, and \textit{ii}.) by estimating the perceptual distortion in the adversarial examples using LPIPS, a normalized distance measure based on deep network activations~\cite{DBLP:journals/corr/abs-1801-03924}. For the former, the results are already summarized in Table~\ref{tab:sparsity}. For the latter, we compare the adversarial examples with their original inputs using LPIPS on AlexNet.
\begin{table}[!ht]
\caption{LPIPS Perceptual Distance ($\times 10^{-3}$). \textbf{Lower is better.}}
\label{tab:lpips}
\vskip 0.05in
\begin{center}
\begin{small}
\begin{sc}
\begin{tabular}{
    >{\arraybackslash}m{0.2cm}|
    >{\arraybackslash}m{2.8cm}|
    >{\centering\arraybackslash}m{1.7cm}
    >{\centering\arraybackslash}m{1.7cm}}
\toprule
&Attacks $\downarrow$ &Vanilla & \textit{delta} \\
\midrule
\parbox[t]{10mm}{\multirow{4}{*}{\rotatebox[origin=c]{90}{\textit{CIFAR-10}}}}
&\textit{delta}           & \;\;0.1          & \;\;9.2 \\
&Unc. + \textit{delta}    & \;\;\textbf{0.0} & \;\;\textbf{9.1}\\ 
&ReColor                  & \;\;6.9          & 11.3    \\
&Unc. + ReColor           & \;\;\textbf{1.1} & \textbf{10.9}\\
\midrule
\parbox[t]{10mm}{\multirow{4}{*}{\rotatebox[origin=c]{90}{\textit{STL-10}}}} 
&\textit{delta}           & \;\;7.9          & 69.0         \\
&Unc. + \textit{delta}    & \;\;\textbf{4.1} & \textbf{41.4}\\ 
&ReColor                  & 22.9             & 23.3         \\
&Unc. + ReColor           & \textbf{10.3}    & \textbf{22.6}\\ 
\bottomrule
\end{tabular}
\end{sc}
\end{small}
\end{center}
\vskip -0.15in
\end{table}
\vskip 0.1in
\noindent LPIPS measures the perceptual distortion between any two images by returning values within $[0, 1]$. We sample at random 100 images from CIFAR-10 and STL-10 and perform \textit{delta} and ReColor attacks under their natural threat models and ours against deterministic classifiers to generate adversarial examples. We compute and report the average LPIPS scores in Table~\ref{tab:lpips}. As the results indicate, the adversarial examples retain a greater degree  of similarity with the original inputs. To illustrate this further, we provide in Appendix~\ref{sec:advers_images} examples of images and their adversaries.

\subsection{Transferability}
\vskip 0.1in
\noindent Adversarial examples often transfer across models~\cite{DBLP:journals/corr/PapernotMG16,42503,ianj.goodfellow2014}. That is, the same examples one generates to evade a specific model can evade other models trained for the same task even when their architectures differ. Here, we ask ourselves: \textit{``How well delta and ReColor attacks transfer under our threat model?"} Explicitly, we seek to validate whether the adversarial examples we generate against deterministic models trained with PGD transfer to stochastic models also defended with PGD, and vice-versa. As Table~\ref{tab:transferability} indicates, both \textit{delta} and ReColor attacks transfer relatively with our threat model in comparison with their own threat model.
\begin{table}[ht]
\caption{Transferability of \textit{delta} and ReColor under our model.}
\label{tab:transferability}
\vskip 0.05in
\begin{center}
\begin{small}
\begin{sc}
\begin{tabular}{
    >{\arraybackslash}m{0.2cm}
    >{\arraybackslash}m{2.4cm}
    >{\centering\arraybackslash}m{1.6cm}
    >{\centering\arraybackslash}m{0.3cm}
    >{\centering\arraybackslash}m{1.6cm}}
\toprule
&Attacks $\downarrow$        & Deter. & $\leftrightarrows$ & Stoch.\\
\midrule
\parbox[t]{10mm}{\multirow{2}{*}{\rotatebox[origin=c]{90}{\textit{MNIST}}}}
&\textit{delta}              & 54.80         & & \;\;0.53         \\[1ex]
&Unc. + \textit{delta}       & \textbf{58.22}& & \;\;\textbf{0.58}\\[1ex]
\midrule
\parbox[t]{10mm}{\multirow{4}{*}{\rotatebox[origin=c]{90}{\textit{CIFAR-10}}}} 
&\textit{delta}              & 73.38         & & \;\;9.40         \\
&Unc. + \textit{delta}       & \textbf{73.60}& & \textbf{10.04}   \\
&ReColor                     & \textbf{66.46}& & \textbf{38.79}   \\
&Unc. + ReColor              & 63.56         & & 36.19            \\
\bottomrule
\end{tabular}
\end{sc}
\end{small}
\end{center}
\vskip -0.15in
\end{table}

\clearpage
\section{Adversarial Images}
\label{sec:advers_images}
Here, we provide more examples of adversarial images that we construct using as natural inputs images drawn from ImageNet and STL-10 datasets.
\subsection*{ImageNet Adversarial Images}
\begin{figure*}[ht!]
\begin{center}
    \subfloat{{\includegraphics[width=8.1cm]{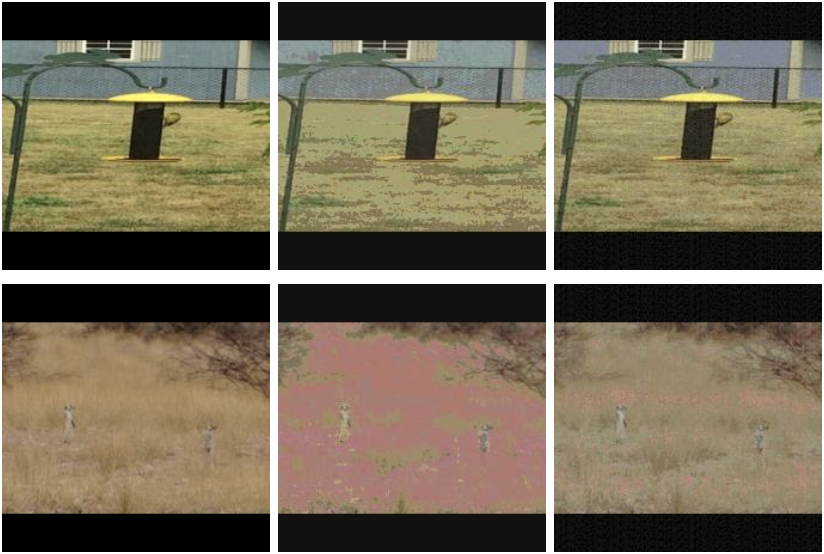} }}%
    \qquad
    \subfloat{{\includegraphics[width=8.1cm]{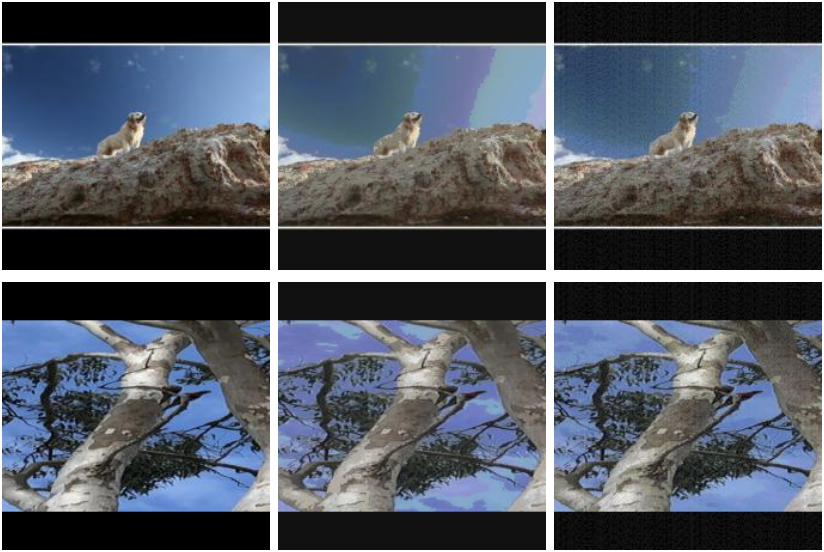} }}%
    \qquad
    \subfloat{{\includegraphics[width=8.1cm]{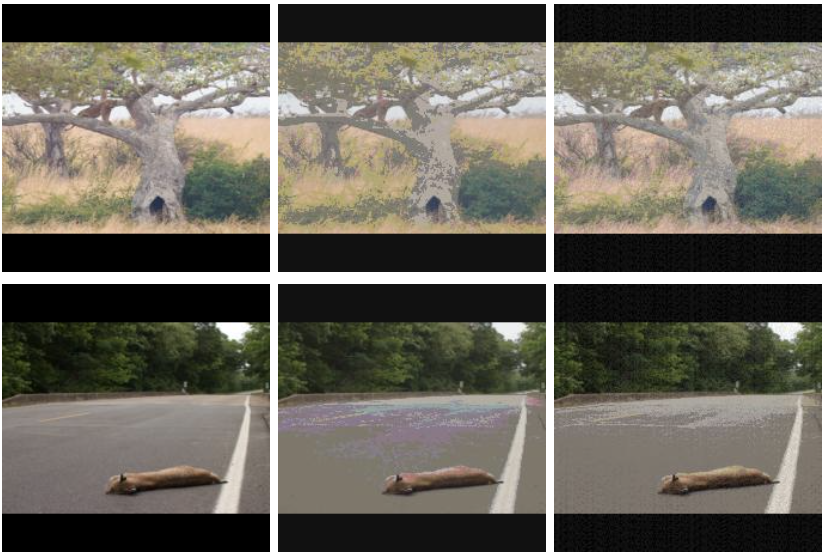} }}%
    \qquad
    \subfloat{{\includegraphics[width=8.1cm]{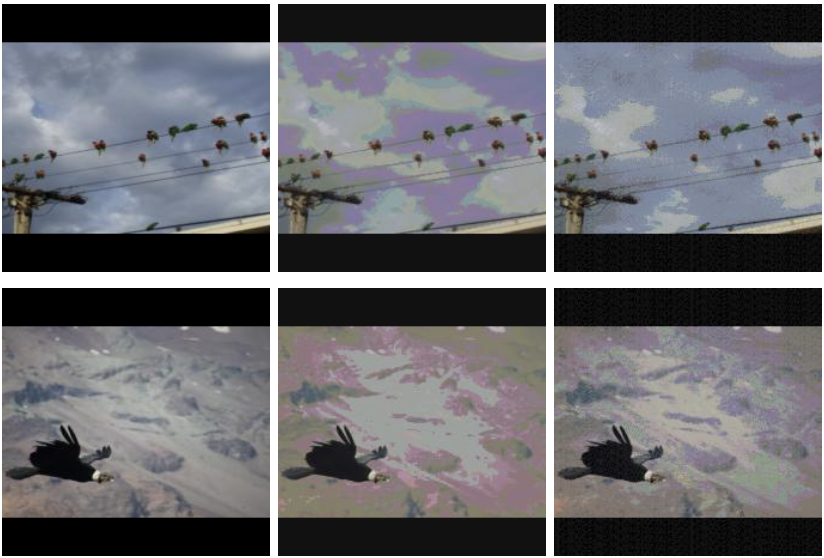} }}%
    \caption{Four groups of ImageNet adversarial examples generated using ReColor under its native functional threat model and under our threat model against an undefended deterministic ResNet-50. From left to right in each row and in each group: original image, adversarial example generated using ReColor, and adversarial example generated using ReColor under our threat model.}%
\end{center}
\end{figure*}
\vspace*{\fill}
\textit{Continued next page ...}
\clearpage
\subsection*{STL-10 Adversarial Images}
\begin{figure*}[h]
\begin{center}
    \subfloat{{\includegraphics[width=8.1cm]{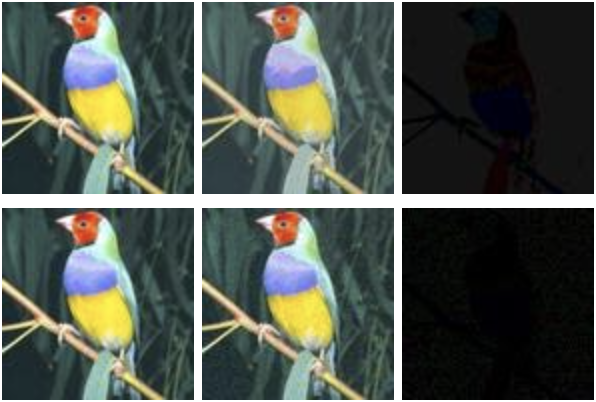} }}%
    \qquad
    \subfloat{{\includegraphics[width=8.1cm]{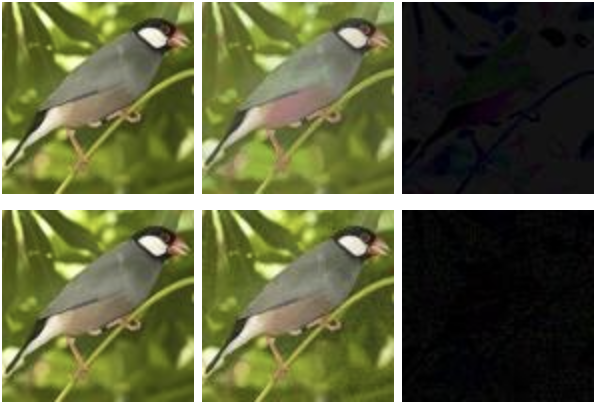} }}%
    \qquad
    \subfloat{{\includegraphics[width=8.1cm]{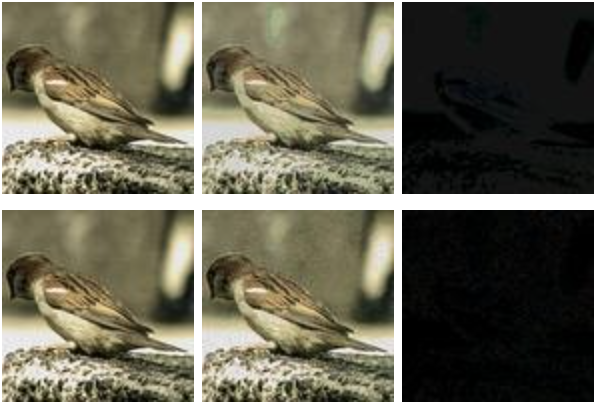} }}%
    \qquad
    \subfloat{{\includegraphics[width=8.1cm]{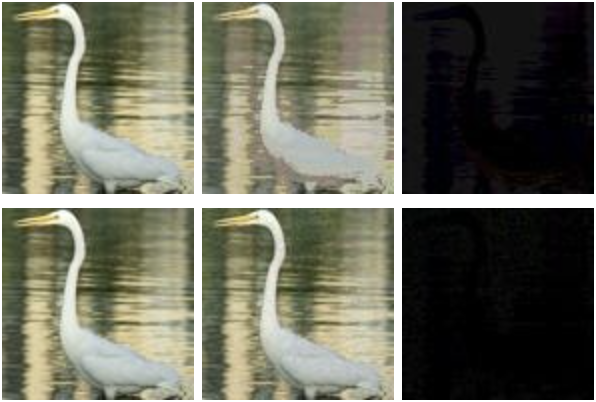} }}%
    \caption{Four groups of STL-10 adversarial examples generated using ReColor under its native functional threat model and under our threat model against an undefended deterministic classifier. From left to right in each group: \textit{top row ---} original image, adversarial example generated using ReColor, and perturbations, \textit{bottom row ---} original image, adversarial example generated using ReColor under our threat model, and perturbations.}%
\end{center}
\end{figure*}

\end{document}